# Polygonal approximation of digital planar curve using novel significant measure

*Mangayarkarasi R*[*,1], Dilip K. Prasad [2]*

[1]Faculty of school of Information Technology & Engineering, Vellore Institute of Technology, Vellore (India) 632014,

[2]School of Computer Science & Engineering, Nanyang Technological University, Singapore 639798,

[*]email: rmangayarkarasi@vit.ac.in

## ABSTRACT

This paper presents an iterative smoothing technique for polygonal approximation of digital image boundary. The technique starts with finest initial segmentation points of a curve. The contribution of initially segmented points towards preserving the original shape of the image boundary is determined by computing the significant measure of every initial segmentation points which is sensitive to sharp turns, which may be missed easily when conventional significant measures are used for detecting dominant points. The proposed method differentiates between the situations when a point on the curve between two points on a curve projects directly upon the line segment or beyond this line segment. It not only identifies these situations, but also computes its significant contribution for these situations differently. This situation-specific treatment allows preservation of points with high curvature even as revised set of dominant points are derived. The experimental results show that the proposed technique competes well with the state of the art techniques.

***Index terms-*** *dominant point, projection position, iterative smoothing, minimal number of points, polygonal approximation.*

## 1  INTRODUCTION

Shape representation and shape classification are efficiently facilitated by polygonal approximation. This approach is popular due to its compact representation and insensitive to noise. These salient features are found useful in many applications [1-8]. The main objective of polygonal approximation is to approximate the shape of a curve using a polygon whose vertices are specified by a subset of points on the curve. These points are referred to as dominant points and are often the points with high curvature. An example is illustrated in Fig.1. A digital curve representing the shape of snowflake is displayed in Fig.1 (a), and its identified dominant points are shown in Fig.1 (b). The anticipated output of polygonal approximation using dominant point can be seen in Fig.1(c). Broadly polygonal/closed curve approximation of a digital planar curve may be cast as min ε problem or min ≠ problem. In min ε problem, the techniques derive polygonal approximation with specified number of line segments or dominant points. These techniques ensure that the deviation between the curve and the approximate polygon is minimal, condition to the specified number of dominant points. Min # techniques derive polygonal approximation with a specified error. These techniques generate the approximate polygon with minimal number of dominant points while ensuring the measure of closeness is not larger than the specified error. In recent years, there are many dominant points based polygonal approximation techniques were presented in the literature [9-19].

And few older ones can be found in [20-22]. The techniques presented in [9, 10, 12, 20, 21] use reverse polygonization, where instead of detecting the real points the techniques makes a search to detect redundant points and deletes points iteratively. The methods in [11, 15] use breakpoint suppression, where the techniques apply criterion measure on the finest approximated set of points to suppress the redundant points and makes the approximation. The methods in [3,13, 16, 18] present a solution using dynamic programming, where the techniques makes exhaustive search to detect points on curve thereby makes final approximation. The method in [14] makes polygonal approximation by detecting ADSS (Approximate Digital Straight Segment).The method in [17] uses MIP (mixed integer programming) model. The method in [19] uses vertex relocation procedure around neighbors. In this method while approximating the output curve by detecting the dominant point, the technique allows neighbourhood points to become a dominant point provided that new dominant point facilitates in reduction of approximation error. The method in [22] uses split and merge, where the method make a search to find the points with maximum deviation in the splitting stage using the proposed criterion function and merge all the points identified in the splitting stage using the threshold value. Most of the dominant points [9-12] detecting methods use the magnitude of orthogonal projection of a point on the line segments which connect adjacent high curvature points to influence the process of detecting dominant points. The methods in the literature [9-12,14,15,,20,23] do not address the issue where the projection of point lies beyond its candidate line segment, where the situation may be often anticipated during approximation. The techniques which neglect to check this criterion may miss good curvature points which are critical for shape representation. The technique proposed in this paper measures the positions of projections of a point on the curve thereby invokes different metric for computing the significant measure of the dominant points. This practice makes the proposed technique to preserve the original shape of the curve even at very minimal number of dominant points. Such characteristic is very essential for compact representation. And it is very essential for object detection and shape classification applications. Rest of the paper is organized is as follows section 2 presents a brief review of some of the state of the art methods along with an insight of their demerits wherever possible. Section 3 presents the proposed work. Section 4 summarizes the experimental results. Section 5 concludes the paper.

**2. BACKGROUND**

Several polygonal approximation techniques have been proposed in the recent decades. Some of them use various optimization approach [3,13,16,17,18,19] On the other hand, there are other techniques that use local/global geometric features of a curve to influence the process of determining the polygon with minimal number of line segments. [9,10, 11, 12, 23,24,25,26], and these techniques prove its competence against many real time datasets. Among these this section briefly analyse some of the bench mark techniques.

Prasad [23] proposed a non-parametric framework to detect points of high curvature. The framework uses the maximum deviation incurred between pixels from a digitized boundary as an upper bound, to make approximation. The authors proved that the analytical bound can be incorporated by dominant point detection framework to get rid of specification in terms of the tolerable error (for min # approaches) or the number of points (for min $\epsilon$ approaches). The authors established the robustness of their framework against scaling invariance as well as noise tolerance. However, there are applications in which the curve needs to be approximated using a specified number of dominant points, which is not possible through this framework. Though the approximation bounded below to digitization value, points detected on the curve seem to be redundant for human visual perception. Prasad [24] used metrics such as precision and reliability as measures to fit the polygon edges. Depending upon the threshold values for these measures, the technique produces coarser or finer approximation. Thus, this technique can flexibly control the degree of smoothness required for an application. And also the paper suggests some performance metrics to quantify the techniques. Parvez [19] obtained the digital boundary using contour extraction techniques. The objective of the method was to produce approximate polygon with minimal error possible. To attain this goal, the method relaxes the criteria that dominant points need not to be on the contour. The techniques computes neighbourhood points for every point $p_i$ on the contour $C_d$ and introduce a new point on the contour provided its presence should reduce the approximation error. The neighbourhood points are not the ones computed using 4 connected graph or 8 connected graph, instead the technique adaptively defines the width for every point on the curve, thereby it obtains the neighbourhood points. Fernandez [25] produced symmetric approximation for symmetric contours. The technique obtains first initial point $p_1$ as the farthest in terms of distance from the centroid of the curve. The next point $p_2$ is the farthest to $p_1$. The method proceeds to find point $p_4$ which is farthest from $p_2$ and point $p_3$ which is farthest from $p_1$. Likewise the technique obtains the all possible line segments such as $\{p_1, p_2\}$, $\{p_3, p_4\}$, until the maximum deviation from the curve does not exceed a threshold value which constitutes the boundary point set. The authors demonstrate that their method of choosing initial points ensures symmetricity. The technique then identifies all possible candidate points ($q_1, q_2, ..., q_m$) from the boundary point set between every two initial points and computes a significant value along with by ensuring symmetry property. Additionally, the technique presents various thresholding methods to normalize the significant values of the boundary points. Though the technique produces symmetric approximation for symmetric curve, it did not establish geometric invariance. And in real time data sets, most of the cases the points are always distributed asymmetrically on the planar curve. The main objectives of this paper are i) present a framework which considers the projection position of a point and thereby invokes the proper criterion measure to compute the contribution. ii) Produce output polygon without missing significant points. iii) Produce polygon with minimal possible number of points. iv) present a technique which is reasonably strong enough against rotation invariance. These objectives are achieved and demonstrated through experimentations of the proposed technique using bench marking data sets.

**3 PROPOSED WORK**

### 3.1.1 Problem formulation

The problem formulation is as follows. Let $C_d = \{p_1,p_2,….p_n\}$ where $p_i = (x_i,y_i)$ is a digital curve consisting of *n* points in clockwise direction in the discrete 2-dimensional space. Such curves are the one extracted from the boundaries of the digital images using contour detection or edge detection methods. The coordinates of these *n* points are integers since these points are extracted from the digital boundary. The objective of polygonal approximation of $C_d$ is to derive a subset $D = \{p_1, p_2, …, p_m\}$ from the super set of $C_d$, subject to the condition the polygon formed by the elements of *D* should represents the shape of the original curve. The technique starts with any three consecutive points $p_i$, $p_j$ and $p_k$ on the curve $C_d$, to detect the collinearity of these points ($p_i$, $p_j$, $p_k$), the distance measured from a point point $p_j$ to the line segment connecting $p_i$ and $p_k$. The method shall conclude the three points are collinear, provided the measured distance very minimal. On the other side, the method shall concludes non-collinearity, provided the measured distance is not very minimal and thus $p_j$ become an element of *D*. Thereby, the polygonal approximation techniques finds all the elements of *D*. With this problem formulation, our paper focuses on the choice of the significant measure metric. Conventionally the distance metric is the length of the line dropped from the point $p_j$ on the line segment $p_ip_k$. This is being referred to as the perpendicular distance. This metric is generally good for smooth curves, but in some cases (explained later) it may miss significant points and rejects sharp turn, which are essential in shape representation applications. Dunham [27] makes initial approximation using distance to a line segment. Ramaiah [28] use distance to a line segment as a measure to make polygonal approximation but the metric used in the technique to compute deviation is capable of preserving sharp turnings but fails to preserve the original shape of digital curve. Apart from the criterion measure proposed in any technique the methodology is also an important factor to produce the output polygon without compromising its actual shape. This implies that the used metric in [28] is unsuitable for iterative smoothing. The framework proposed in this paper automatically chooses the suitable significant measure metric based on the candidate point projection, as explained next.

### 3.1.2 Proposed technique

In this section, we present our proposed method to make polygonal approximation of $C_d$. The initial segmentation points are obtained using Freeman chain code [28], such as given in Algorithm 1. These initially segmented points are referred as initial set of dominant points. Example of initial segmentation for the snowflake curve is shown in Fig. 1(a) is given in Fig. 1(b) where the dominant points are highlighted in bold markers and the final approximated curve is given in Fig. 1(c).

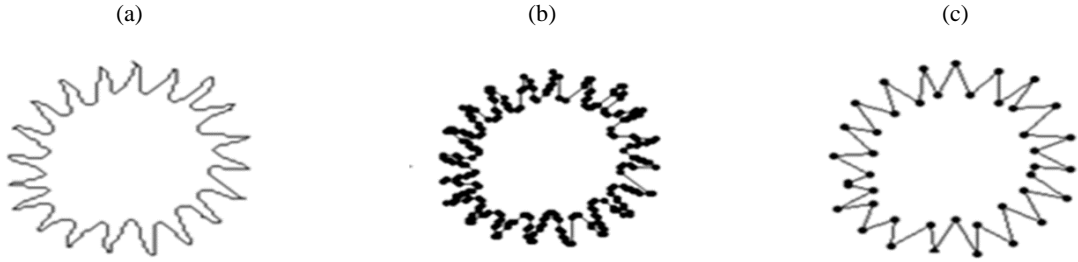

(a)  (b)  (c)

**Fig.1: a)A digital curve representing the shape of a snowflake, b) Initial set of dominant points, c)Suitable polygonal approximation are shown here.**

To compute the significant measure of every initial dominant point $s_k$, the proposed method use the following steps. Consider the scenario in Fig. 2(a) where namely, $s_{k-1}$, $s_k$ and $s_{k+1}$ are three dominant points on the curve with the following traversing sequence: $s_{k-1} \rightarrow s_k \rightarrow s_{k+1}$. It may be interpreted as these three points are collinear by assuming the projections of a point $s_k$ is lies on the line segment which connects $(s_{k-1} s_{k+1})$. As a consequence the approximation technique [9-12,14-15,20,23,39-40] may decide to drop $s_k$. In this scenario the projection of a point ($s_k$) not lies between the its candidate line segment $(s_{k-1} s_{k+1})$. The Fig. 3 shows the various anticipated position for possible projection of a dominant point ($s_k$) on the x-y plane. The proposed metric detects the position of projection. In order to predict the position of a projection, the proposed technique uses the following steps. Translate the line segment connecting $s_{k-1}$ and $s_{k+1}$ so that the point $S_i$ coincides with the origin of the xy coordinate system and measure the amount of angle produced by the translated line segment with the x axis. In order to align the translated line segment with the x axis, rotate the line segment with a computed amount angle. The actual x-y coordinates system and new transformed coordinate systems are displayed in Fig. 2(a) and Fig. 2(b). In the next step by checking transformed x coordinate of $s_k$' the method chooses metric to compute the significant measure. If the x coordinate of $s_k$' is less than 0 then the significant measure $sig(s_k)$ is computed using eqn. (1).(see Fig. 3(a)) if $x_k$' of $s_k$' lies between 0 and the x coordinate of $s_i$ then the significant measure is computed using eqn. (2) (see Fig. 3(b). If the $x_k{}'$ value is greater than $x_j{}'$ of $s_{k+1}$ then the significant measure of $s_k$ is computed using eqn. (3) (See Fig. 3(c)).

$$sig(s_k) = \sum_{k=s_{k-1}}^{s_{k+1}} \sqrt{(s_{x_k} - s_{x_{k-1}})^2 + (s_{y_k} - s_{y_{k-1}})^2} \quad (1)$$

$$sig(s_k) = \sum_{k=s_{k-1}}^{s_{k+1}} |S_{y_{k'}}| \quad (2)$$

$$sig(s_k) = \sum_{k=s_{k-1}}^{s_{k+1}} \sqrt{(s_{x_k} - s_{x_{k+1}})^2 + (s_{y_k} - s_{y_{k+1}})^2} \quad (3)$$

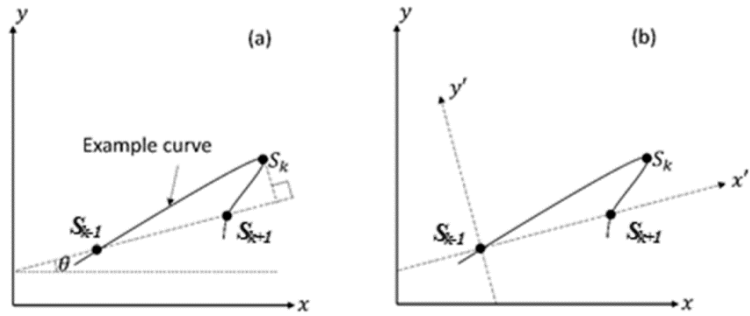

**Fig. 2:** Demonstration of the coordinate transform performed for the proposed self-adaptive significant measure computing metric for dominant point detection. (a) An example curve in the original x-y coordinate system is shown. (b) The transformed x'-y' coordinate system is shown in addition to the original x-y system

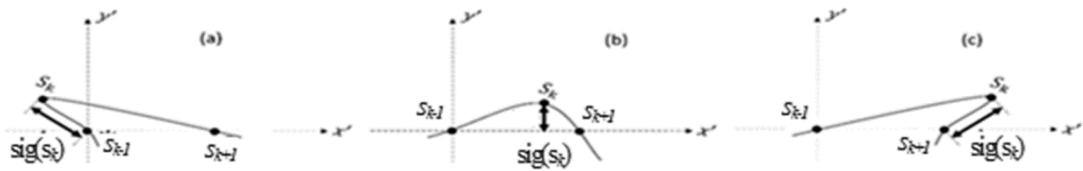

**Fig.3:** Demonstration of computation of significant measure of the point $s_k$ from the line segment $s_{k-1}$ $s_{k+1}$.

In all the three equations (eqn. (1),eqn.(2) and eqn. (3)), $k$ range is, $k-1<=k<=k+1$.(Note: the accent sign indicates the coordinates in the transformed coordinate system). While computing the significant measure associated with a dominant point let us say $s_k$, the significant measure of every non-dominant point/ boundary point lies between its candidate line segment are accumulated to define the significance measure of $s_k$. These steps are repeated for each dominant point in the initial set, before making the decision to remove redundant dominant points in the next step. After measuring the significant measure of all initial dominant points, the proposed method removes the dominant point with minimal significant measure. If more than one dominant point has the same minimal significant measure, the dominant point appearing first in the order of sequence is removed. The steps to remove the dominant point and producing the final output polygon are given in Algorithm 2.

**Algorithm 1: CIDP (Compute initial set of dominant points)**

```
Input:   The input are the coordinates of the boundary points.
         C_d = p_i (x_i,y_i), i=1,2,3…..n; n boundary points.
Output: The outputs are the curve indices of initial dominant points.
Begin
Case 1: i=0
        If (x(0)-x(n-1) != x(1)-x(0)) or ((y(1)-y(0) != y(0)-y(n-1)) then
        D[0]= 0;
Case 2: i=n-1
        If (x(n-1)-x(n-2) != x(0)-x(n-1)) or (y(n-1)-y(n-2) != y(0)-y(n-1))
        D[j]=i;
Default:
        While (i<n-1)
        If (x(i)-x(i-1) != x(i+1)-x(i)) or (y(i+1)-y(i) != y(i)-y(i-1))
          D[j] = i
End.
```

**Algorithm 2: Polygonal approximation by computing the significant measure of IDP**

```
Input : Digital curve C_d
      :Number of dominant points (m) in the output polygon
Output: Output polygon with the specified number of dominant points (k)
Begin
Step1 : Invoke the function CIDP
Step2 : Compute significant measure associated with all initial dominant points (s_k's)
Step 3: Repeat
```

| | | |
|---|---|---|
| | i) Identify the dominant point $s_k$ with minimal significant measure in $C_d$ | |
| | ii) Remove the dominant point $s_k$ and recalculate the significant measure of at $s_{k-1}$ and $s_{k+1}$ | |
| | iii) Compute the performance measures with the available dominant points | |
| | Until (No.of.DPs == $k$) | |
| **End** | | |

## 4. Experimental results

The proposed technique is tested on a variety of challenging curves to demonstrate its efficiency. The results are presented for two experiment sets. The experiment set 1 consists of synthetic curves usually used in the literature [9, 11, 16, 19, 24, 25, 29, 30, 31, 32, 33,34,35,36]. In the experiment 2, the proposed method is tested extensively with images in MPEG dataset [37].We first present the quality assessment metrics for polygonal approximation of digital curves. Then, we present the results on the two experimental sets. Additionally, we include one experiment to demonstrate geometric invariance of the proposed technique.

4.1.1 Quality assessment

The best method to assess output of polygonal approximation is visual perception. Thus, we include extensive qualitative results. Moreover, we include quantitative performance measures as well for comparison of the performance of the tested methods, including the proposed technique. This paper considers the following metrics to measure the goodness of the results: i) Compression ratio (CR), ii) Integral square error (ISE), iii) Figure of merit (FOM),iv) Weighted sum of square errors (WE),v) Modified version of WE (WE2).Details of these metrics are provided in Table 1. These metrics are taken from [9-11, 15, 19, 31, 34]. The readers interested in them are encouraged to read these articles and the references therein.

**Table 1: Quality assessment metrics for comparing polygonal approximation methods.**

| Metric | Indicator of goodness | Mathematical representation |
|---|---|---|
| CR | Larger is better | $CR = \frac{n}{m}$, where $n$ is the number of points in the initial segmentation while $m$ is the number of dominant points in the final polygonal approximation. |
| ISE | Smaller is better | $ISE = \sum_{k=1}^{n} e_k$, where $e_k$ is the perpendicular distance of a point $p_k$ on the original digital curve from the nearest line segment on the polygonal approximation. |
| FOM | Larger is better | $FOM = \frac{CR}{ISE}$ |
| WE | Smaller is better | $WE = \frac{ISE}{CR}$ |
| WE2 | Smaller is better | $WE2 = \frac{ISE}{CR^2}$ |

4.1.1 Experiment set 1

The quantitative performance measure for the synthetic curves chromosome, leaf, semicircle and infinity in experiment set 1 are given in Table 2. The visual shots are shown in Fig. 4-7. The methods in [16,17,18,19,32,34,35] presents an optimal solutions for the polygonal approximation. The proposed method output is close to optimal solution for all the curves and further supports reduction of the number of dominant points while retaining the shape information of the curve. The Table 2 summarizes the results from various articles [9,11, 15, 16, 17, 18, 19,23, 24, 26, 29, 30, 31, 32, 33, 34,35,36] for the given input synthetic curves. For the chromosome curve display using 15 amount of dominant points the proposed technique produces a low value for ISE than the method in [30,31,32]. The snapshot of chromosome curve at 6 number of points using the proposed method as well as by the methods [9,23,24] snapshots can be found in Fig. 4. For the leaf curve, where the output curve at 21 number of dominant points, the proposed method produces the low value for ISE than [11,24,32] (in turn FOM value is high which is appreciable) and high value than [19]. The snapshot for leaf output curve produced by the proposed method along with some of the state of the art methods results are displayed in Fig. 5. The output description for the semicircle shaped curve is as follows. While giving an attempt to display the semicircle curve at 17 number of dominant points the proposed method results are better than [32,35] in terms of ISE ,WE and FOM.  Then the output polygon using 14 number of dominant points proposed method results in terms of ISE WE an FOM are better than [18,34], and comparable with [17,36] . (Whereas the method [18,35] use genetic algorithms)

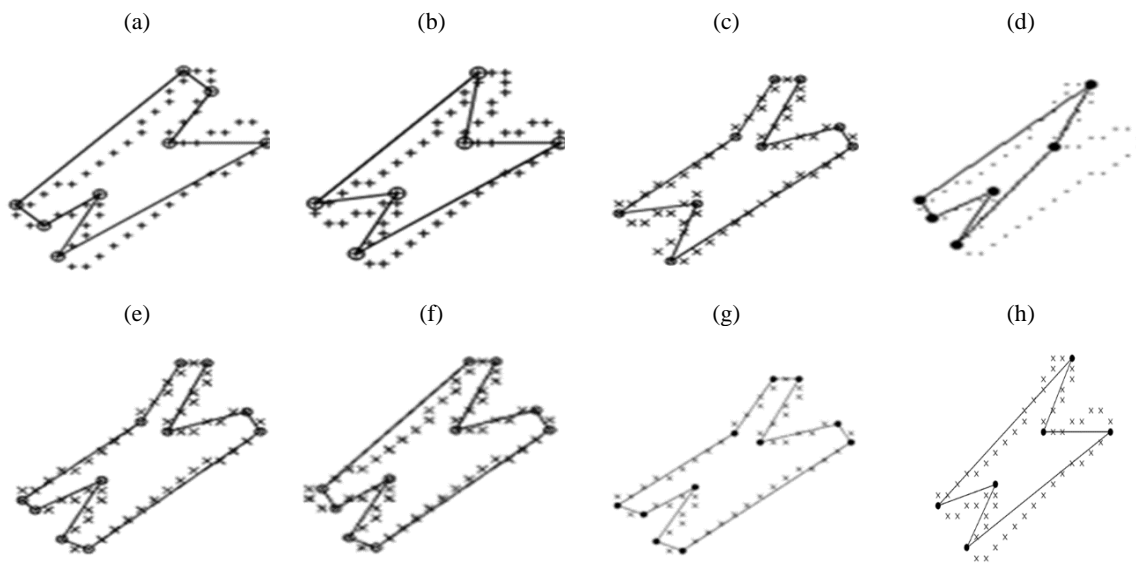

**Fig. 4:** Polygonal approximation of chromosome curve at varying amount of dominant points. a) RDP2[24] at 11 DPs, b) RDP3[24] at 6 DPs, c) Masood[9] at 9 DPs, d) Masood[9] at 6 DPs, e) Prasad [23]Masood_opt at 11 DPs, f) Prasad[23] Carmona_opt at 10 DPs, g) Proposed method at 11 DPs, h) Proposed method at 6 DPs.

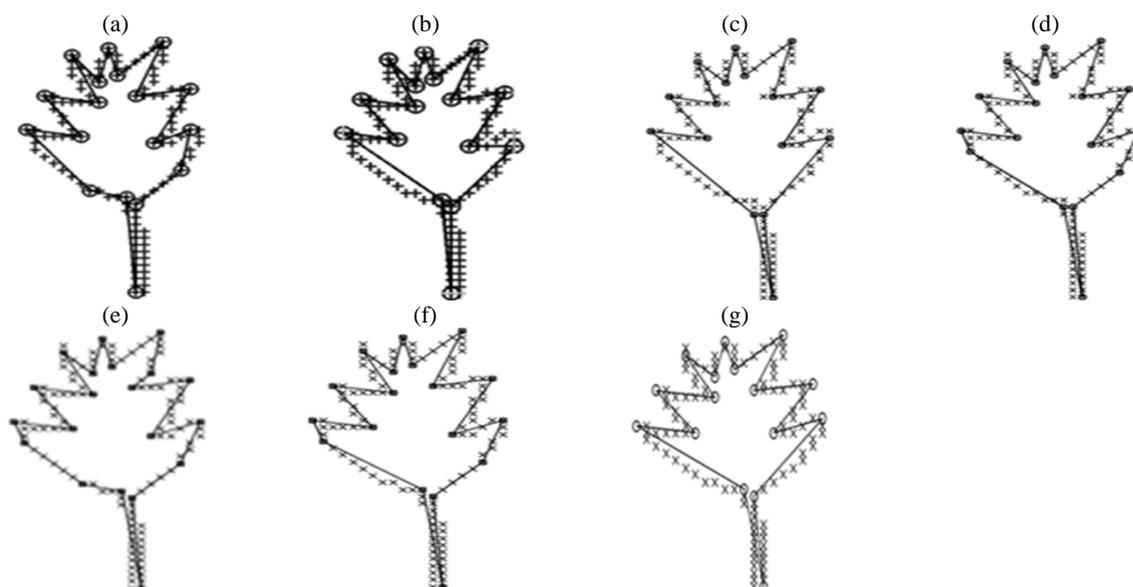

**Fig. 5:** Polygonal approximation of leaf curve at varying amount of points. a) Prasad[24]PRO 0.6 at 18 DPs, b) Prasad[24]RDP2 at 16 DPs, c) Masood[9] at 16 DPs, d) Prasad[23] Masood_opt at 18 DPs, e) Carmona[11] at 20 DPs, f) Prasad[23] Carmona_opt at 18 DPs, g) Proposed method at 16 DPs

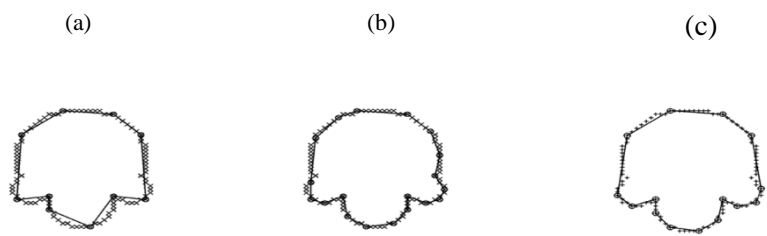

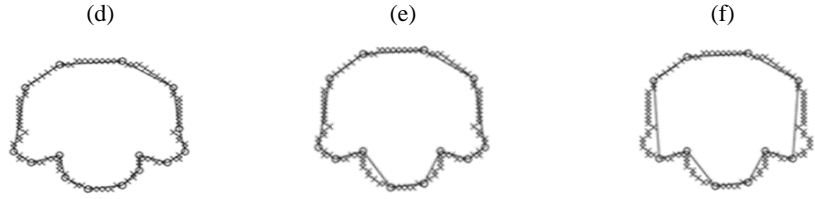

**Fig. 6: Polygonal approximation of semicircle curve at varying amount of points. a) Carmona[11] at 10 DPs, b) Masood[9] at 19 DPs, c) Prasad[24]PRO 0.6 at 15 DPs, d) Proposed method at 16DPs f) Proposed method at 12DPs, f)Proposed method at 10DPs**

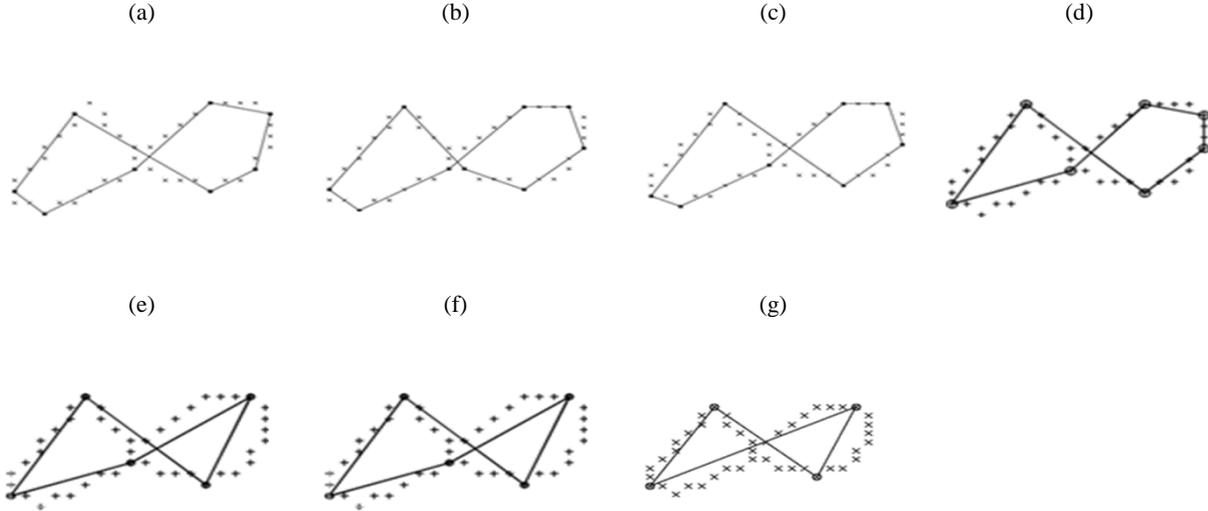

**Fig. 7: Polygonal approximation of infinity curve at varying amount of DPs. a) Masood [9] at 8 DPs, b)Prasad[23] Masood _opt at 9 DPs, c) Carmona[11] at 8 DPs, d) Carmona [11] at 7 DPs, e)Prasad[24] PRO 1.0 at 7 DPs, f)Prasad[24]_RDP 3 at 5 DPs, g) Proposed method at 6 and 4 DPs.**

Table 2: Comparative results of synthetic contour (Chromosome, Leaf, Semicircle, Infinity)

| Contour | Methods | m | CR | ISE | WE | FOM |
|---|---|---|---|---|---|---|
| **Chromosome** | Teh and Chin [30] | 15 | 4.00 | 7.20 | 1.80 | 0.56 |
| $n = 60$ | Wu[31] | 15 | 4.00 | 7.20 | 1.80 | 0.56 |
| | Masood [9] | 12 | 5.00 | 7.76 | 1.55 | 0.64 |
| | Carmona et al [11] | 11 | 5.45 | 14.49 | 2.66 | 0.38 |
| | Parvez [32] | 10 | 6.00 | 14.34 | 2.39 | 0.42 |
| | Madrid et al[26] | 12 | 5.00 | 5.82 | 1.16 | 0.86 |
| | Nguyen and debled-Rennesson [33] | 25 | 3.33 | 4.06 | 1.22 | 0.82 |
| | Nguyen and debled-Rennesson [33] | 15 | 4 | 5.69 | 1.42 | 0.70 |
| | Parvez [19] | 11 | 5.45 | 7.09 | 1.30 | 0.77 |
| | Aguilera et al.[17] | 10 | 6.00 | 8.07 | 1.35 | 0.74 |
| | Lie et al [18] | 14 | 4.29 | 7.58 | 1.77 | 0.57 |
| | Lie et al [18] | 12 | 5.00 | 7.96 | 1.59 | 0.63 |
| | PRO0.6 [24] | 11 | 5.45 | 11.00 | 2.02 | 0.50 |
| | RDP2[24] | 8 | 7.50 | 59.99 | 8.00 | 0.13 |
| | RDP3[24] | 6 | 10.00 | 91.18 | 9.12 | 0.11 |
| | Proposed | 15 | 4.00 | 4.87 | 1.22 | 0.82 |
| | **Proposed** | **6** | **10.00** | **45.49** | **4.55** | **0.22** |
| Leaf | Teh and Chin [30] | 29 | 4.14 | 14.96 | 3.61 | 0.28 |
| $n = 120$ | Wu [31] | 24 | 5.00 | 15.93 | 3.19 | 0.31 |
| | Marji and siy [15] | 17 | 7.06 | 28.67 | 4.06 | 0.25 |
| | Carmona et al [11] | 21 | 5.71 | 17.97 | 3.15 | 0.32 |
| | Parvez [32] | 21 | 5.71 | 13.82 | 2.42 | 0.41 |
| | Parvez [19] | 21 | 5.71 | 11.98 | 2.10 | 0.48 |
| | Nguyen and debled-Rennesson [33] | 33 | 3.64 | 5.56 | 1.53 | 0.65 |
| | Backes and Bruno [34] | 20 | 6.00 | 14.1 | 2.35 | 0.43 |
| | Wang et al [16] | 20 | 6.00 | 13.9 | 2.32 | 0.43 |
| | Madrid et al[26] | 22 | 5.45 | 11.16 | 2.05 | 0.49 |
| | PRO0.6[24] | 21 | 5.71 | 21.70 | 3.80 | 0.26 |
| | PRO1.0[24] | 18 | 6.67 | 36.70 | 5.50 | 0.18 |
| | RDP1[24] | 22 | 5.45 | 19.17 | 3.51 | 0.28 |
| | RDP2[24] | 16 | 7.50 | 65.46 | 8.73 | 0.11 |
| | Proposed | 21 | 5.71 | 13.25 | 2.32 | 0.43 |
| | **Proposed** | **16** | **7.50** | **44.52** | **5.94** | **0.17** |
| Semicircle $n = 102$ | Teh and Chin[30] | 22 | 4.64 | 20.61 | 4.44 | 0.23 |
| | Yin [35] | 17 | 6.00 | 19.78 | 3.30 | 0.30 |
| | Salotti [36] | 14 | 7.29 | 17.39 | 2.39 | 0.42 |
| | Wu [31] | 27 | 3.78 | 9.01 | 2.38 | 0.42 |
| | Marji and Siy[15] | 15 | 6.80 | 22.70 | 3.34 | 0.30 |
| | Masood[9] | 21 | 4.86 | 9.82 | 2.02 | 0.49 |
| | Carmona et al [11] | 26 | 3.92 | 4.91 | 1.25 | 0.80 |
| | Parvez [32] | 17 | 6.00 | 19.02 | 3.17 | 0.32 |
| | Nguyen and debled-Rennesson [33] | 25 | 4.08 | 5.42 | 1.33 | 0.75 |
| | Backes and Bruno [34] | 14 | 7.29 | 19.80 | 2.72 | 0.37 |
| | Wang et al [16] | 15 | 6.80 | 14.30 | 2.10 | 0.48 |
| | Parvez [19] | 15 | 6.80 | 18.22 | 2.68 | 0.37 |
| | Aguilera et al [17] | 14 | 7.29 | 17.39 | 2.39 | 0.42 |
| | Madrid et al[26] | 10 | 10.20 | 40.79 | 4.00 | 0.25 |
| | Lie et al [18] | 14 | 7.29 | 29.30 | 4.02 | 0.25 |
| | PRO 0.6 [24] | 18 | 5.67 | 18.12 | 3.20 | 0.31 |
| | Proposed | 18 | 5.67 | 15.45 | 2.72 | 0.37 |
| | Proposed | 17 | 6.00 | 16.59 | 2.76 | 0.36 |
| | Proposed | 14 | 7.29 | 17.73 | 2.43 | 0.41 |
| | **Proposed** | **12** | **8.50** | **40.62** | **4.78** | **0.21** |
| Infinity $n = 45$ | Teh and Chin [30] | 13 | 3.46 | 5.93 | 1.71 | 0.58 |
| | Wu [31] | 13 | 3.46 | 5.78 | 1.67 | 0.60 |
| | Masood [9] | 11 | 4.09 | 2.90 | 0.71 | 1.41 |
| | Carmona et al [11] | 10 | 4.50 | 5.29 | 1.18 | 0.85 |
| | Parvez [32] | 9 | 5.00 | 7.35 | 1.47 | 0.68 |
| | Parvez [19] | 7 | 6.43 | 7.69 | 1.20 | 0.84 |
| | Madrid et al[26] | 10 | 4.50 | 6.40 | 1.42 | 0.70 |
| | PRO0.6[24] | 9 | 5.00 | 6.29 | 1.26 | 0.79 |
| | PRO1.0[24] | 7 | 5.63 | 19.94 | 3.54 | 0.28 |
| | RDP1[24] | 9 | 5.00 | 6.67 | 1.33 | 0.75 |
| | RDP2[24] | 7 | 6.43 | 19.94 | 3.10 | 0.32 |
| | RDP3[24] | 5 | 9.00 | 53.82 | 5.98 | 0.17 |
| | Masood [9] | 8 | 5.63 | 10.24 | 1.82 | 0.55 |
| | Carmona et al [11] | 6 | 7.50 | 31.68 | 4.22 | 0.24 |
| | Proposed | 10 | 4.50 | 4.44 | 0.99 | 1.01 |
| | **Proposed** | **5** | **9.00** | **35.61** | **3.96** | **0.25** |

The snapshot of the semicircle curve using varying amount of dominant points by the proposed method along with some other counter-part methods are showed in Fig. 6. The final synthetic curve for this experiment set is a curve which intersects itself i.e infinity shaped curve. In the attempt of producing the output curve using 10 number of points the proposed produce the minimal possible error than [11,26]. And also the summarized results reveal that the proposed method output is better than [9,11,19,24,31,26] in terms of ISE, WE and FOM. The graphic shots for the same can be found in Fig. 7. According to human visual perception four points sufficient enough to represents the infinity curve, please see the Fig. 7(g). On the outset, it is perceived that the proposed technique gives the best or second best ISE values for all the cases. This indicates competitiveness of the proposed technique.

### 4.1.2 Experiment set 2

In this section the performance of the proposed methods has been demonstrated using image in MPEG database [37]. Fernandez [25] presents technique to produce output polygon from a given digital boundary. Authors in [25] demonstrated the efficiency of their method by comparing their results with method [23] which is capable of producing output polygon in non-parametric mode. So the better counter-part method to compare the proposed method is the one proposed in [25]. The Table 3 summarizes the results of the proposed method along with the results claimed as the best in [25] for the contours in MPEG database [37]. For the Bell-7 contour the snapshot at 23, 22, 20 and 7 number of dominant points, the proposed method produces a less approximation error in terms of ISE WE WE2 than others mentioned in [9,11,23,25]. Especially the output approximation at 7 DPs the proposed method and Rosin [38] method produces the curve with the mandatory points compared to others, but the proposed method produces minimal error measure than Rosin[38], the output can be found in Fig. 8(c) and Fig. 8(h).

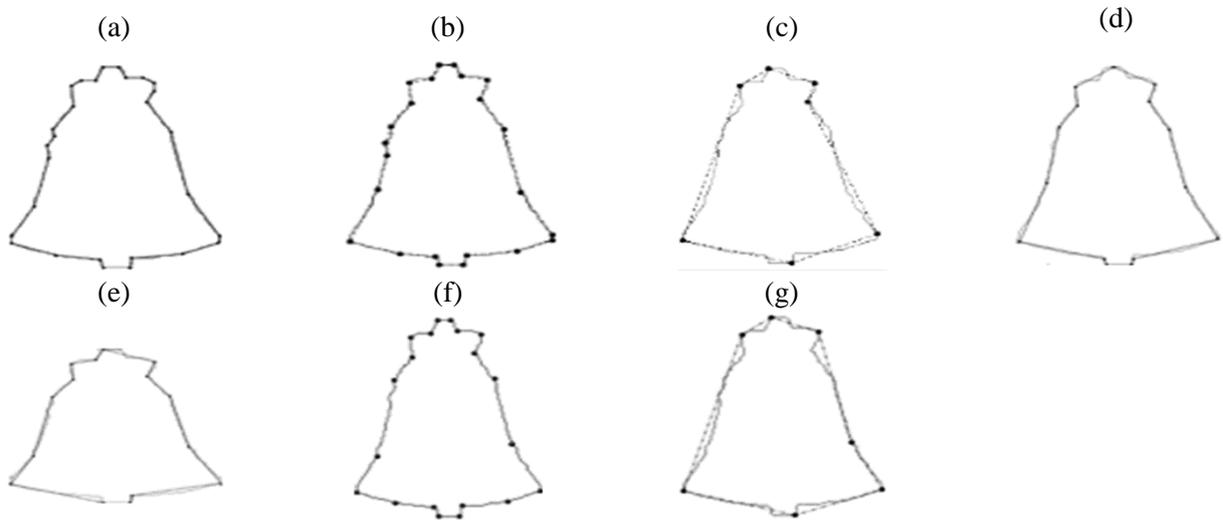

**Fig.8:The output approximation for the bell-7 contour by various methods a)Prasad[23]RDP at 28 DPs b) Fernandez [25] at 23 DPs c)Rosin[38] at 7 DPs,d)Masood[9] at 15 DPs,e)Carmona [11] at 15 DPs f)Proposed method 20 DPs, g)Proposed method at 7 DPs**

For the octopus-14 contour the proposed efficiently produces the output curve with minimal deviation from the original curve compare to others. By observing Fig. 9(e) the proposed produces an outlying approximation which is visibly excellent than [25]. In order to support this claim the output curve for octopus-14 can be found in Fig. 9 along with results of [11,23,25]. When the input is the ray-17 contour, at 14 number of DPs the new proposal produces minimal error than the results of [9,38], then for the same curve at 35 DPs the results are good than [25] in terms of ISE,WE,WE2.The graphic shots of the proposed method along with [11,23,25,38] can be found in Fig. 10. When th input for the proposed method is chicken-5 curve, the proposed method approximation error measures are compared with results produced by the techniques in[9,11,23,25,39,40], by using all the quantitative metric the proposed work produces the output curve with minimal error possible, and the visual snapshots are shown in Fig 11. For device6-9, bell-10 and butterfly-13, the proposed method results are compared with the results in[9,11,23,25,39,40], it is been conceived that produces the minimal error (ISE ,WE) than the error produces by the methods in [9,11,23,25]. Except for the bell-10 curve, Prasad[23] RDP_opt produces the minimum error than the proposed method at 110 dominant points. The output curve for device6-9 can be found in Fig. 12. Then finally for the truck-07 curve, the results of the proposed method at 40,12 and 11 dominant points are compared with the results of [9,11,23,25]. In all iterations against the mentioned dominant points the proposed method outperforms well than others. Especially output curve at 11 dominant points the proposed method efficiently chooses the good curvature points in such a way the output curve do not deviate much than the original input curve. (plz see the snapshot at Fig. 13 (a),(b) with (g)).

**Table 3. Comparative results for the MPEG-database contours**

| Contour | Methods | k | CR | ISE | WE | $WE_2$ |
|---|---|---|---|---|---|---|
| Bell-7 | Fernandez [25] | 23 | 17.65 | 165.14 | 9.35 | 0.53 |
| $n = 407$ | Fernandez [25] | 22 | 18.45 | 200.93 | 10.89 | 0.59 |
|  | Fernandez [25] | 20 | 20.3 | 255.083 | 12.56 | 0.61 |
|  | Rosin [38] | 7 | 58 | 2186.6 | 37.7 | 0.65 |
|  | Masood [9] | 20 | 20.35 | 408.08 | 20.5 | 0.98 |
|  | Carmona [11] | 23 | 17.69 | 332.563 | 8.84 | 0.23 |
|  | Prasad[23]RDP | 28 | 14.53 | 97.60 | 6.71 | 0.46 |
|  | Proposed | 22 | 18.5 | 176.54 | 9.54 | 0.51 |
|  | Proposed | 20 | 20.35 | 210.16 | 10.32 | 0.50 |
|  | **Proposed** | **7** | **58.14** | **453.91** | **7.80** | **0.13** |
| Octopus-14 | Fernandez [25] | 79 | 15.33 | 236.62 | 15.44 | 1.00 |
| $n = 1211$ | Fernandez [25] | 55 | 22.02 | 1270.17 | 57.69 | 2.62 |
|  | Fernandez [25] | 50 | 24.22 | 1847.81 | 76.29 | 3.15 |
|  | Rosin[38] | 43 | 28.16 | 2617.37 | 92.94 | 3.30 |
|  | Masood [9] | 201 | 6.02 | 9268.43 | 1538.36 | 255.75 |

| | Method | | | | | |
|---|---|---|---|---|---|---|
| | Prasad[23]RDP | 55 | 22.01 | 392.15 | 17.81 | 0.80 |
| | Proposed | 79 | 15.33 | 212.00 | 13.83 | 0.90 |
| | **Proposed** | **43** | **28.16** | **1927.15** | **68.43** | **2.42** |
| Ray-17 | Fernandez [25] | 35 | 19.69 | 240.26 | 12.20 | 0.62 |
| $n = 689$ | Fernandez [25] | 28 | 24.61 | 660.00 | 26.82 | 1.09 |
| | Fernandez [25] | 24 | 28.71 | 1152.83 | 40.16 | 1.40 |
| | Rosin[38] | 14 | 49.21 | 6999.71 | 142.23 | 2.89 |
| | Masood[9] | 24 | 28.71 | 749.01 | 26.09 | 0.91 |
| | Masood[9] | 14 | 49.21 | 8627.89 | 175.31 | 3.56 |
| | Prasad[23]RDP | 54 | 12.75 | 342.36 | 26.93 | 2.10 |
| | Proposed | 35 | 19.69 | 208.48 | 10.59 | 0.53 |
| | **Proposed** | **14** | **49.21** | **455.32** | **9.25** | **0.18** |
| Chicken-5 | RDP [39,40] | 255 | 5.35 | 285.54 | 53.38 | 9.98 |
| $n = 1364$ | Masood [9] | 401 | 3.40 | 147.86 | 43.47 | 12.79 |
| | Carmona et al [11] | 134 | 10.18 | 906.52 | 89.06 | 8.74 |
| | Fernandez [25] | 54 | 25.26 | 2424.51 | 95.99 | 3.80 |
| | Prasad[23]RDP | 218 | 6.25 | 782.53 | 125.20 | 20.3 |
| | Proposed | 255 | 5.35 | 275.42 | 51.49 | 9.61 |
| | Proposed | 54 | 25.26 | 1994.15 | 78.95 | 3.12 |
| Device6-9 | RDP [39,40] | 50 | 31.80 | 303.37 | 9.54 | 0.30 |
| $n = 1590$ | Masood [9] | 84 | 18.93 | 189.89 | 10.03 | 0.53 |
| | Carmona [11] | 22 | 72.27 | 3395.17 | 46.98 | 0.65 |
| | Fernandez [25] | 33 | 48.18 | 348.22 | 7.23 | 0.15 |
| | Prasad[23]RDP | 38 | 41.84 | 741.416 | 17.02 | 0.42 |
| | Proposed | 84 | 18.93 | 216.24 | 11.42 | 0.60 |
| | **Proposed** | **22** | **72.27** | **761.58** | **10.54** | **0.14** |
| Bell-10 | RDP [39,40] | 110 | 10.92 | 181.25 | 16.59 | 1.52 |
| $n = 1202$ | Masood [9] | 4 | -- | -- | | 4.95 |
| | Carmona [11] | 104 | 11.78 | 549.52 | 46.64 | 3.96 |
| | Fernandez [25] | 42 | 28.61 | 687.56 | 24.03 | 0.84 |
| | Prasad[23]RDP | 81 | 14.83 | 326.47 | 22.01 | 1.48 |
| | Proposed | 110 | 10.92 | 241.45 | 22.06 | 2.02 |
| | **Proposed** | **42** | **28.61** | **615.77** | **7.98** | **0.75** |
| Truck-07 | RDP [39,40] | 40 | 6.92 | 24.45 | 3.53 | 0.50 |
| $n = 277$ | Masood[9] | 40 | 6.92 | 37.17 | 5.37 | 0.77 |
| | Masood [9] | 11 | 25.18 | 1133.29 | 45.00 | 1.78 |
| | Carmona [11] | 12 | 23.08 | 1132.45 | 49.06 | 2.11 |
| | Fernandez [25] | 40 | 6.92 | 24.15 | 3.48 | 0.50 |
| | Prasad [23]RDP | 33 | 8.39 | 59.17 | 7.05 | 0.84 |
| | Proposed | 12 | 23.08 | 319.24 | 13.83 | 0.59 |
| | **Proposed** | **11** | **25.18** | **318.34** | **12.64** | **0.50** |
| Butterfly-13 | RDP [39,40] | 344 | 5.19 | 383.30 | 73.85 | 14.23 |
| $n = 1786$ | Masood [9] | 525 | 3.40 | 199.06 | 58.54 | 17.22 |
| | Carmona [11] | 171 | 10.44 | 1450.70 | 138.95 | 13.31 |
| | Fernandez [25] | 65 | 27.47 | 2195.88 | 79.93 | 2.91 |
| | Proposed | 525 | 3.40 | 197.58 | 58.11 | 17.09 |
| | **Proposed** | **65** | **27.47** | **2063.91** | **75.13** | **2.73** |

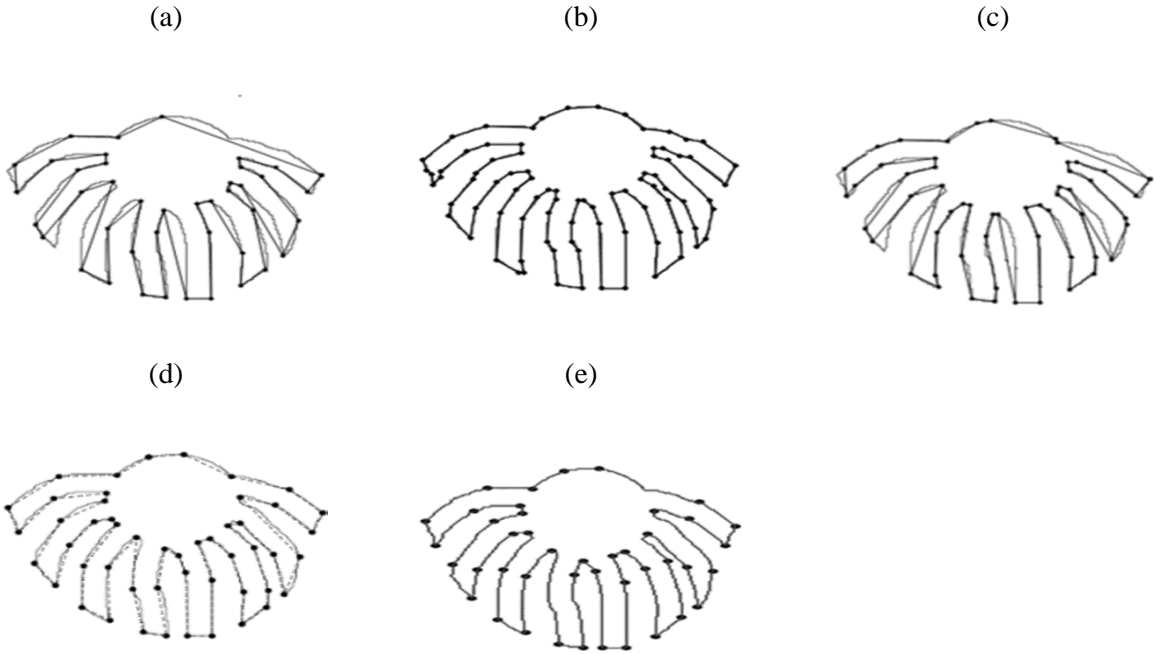

**Fig. 9: The output polygon from octopus-17 by various method a) Carmona [11] at 43 DPS b) Prasad [23]RDP at 55 DPs c) Prasad [23]Carmona_opt d) Fernandez [25] at 43 DPS e) Proposed method at 43 DPs.**

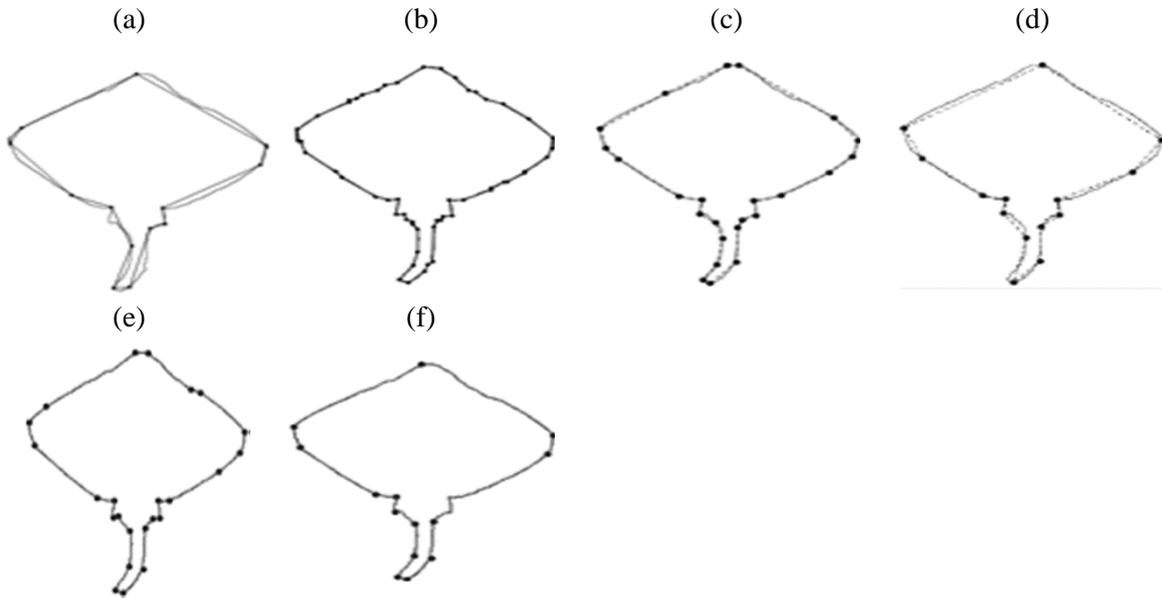

**Fig 10. Output approximated curve for ray-17 contour by various method a)Carmona [11] at 14 DPs b) Prasad [23]RDP_opt at 54 DPs c) Fernandez [25] at 24 DPs d) Rosin[38] at 14 DPs e) Proposed method results at 24 DPs f) Proposed method at 14 DPs**

(a) (b) (c) (d)

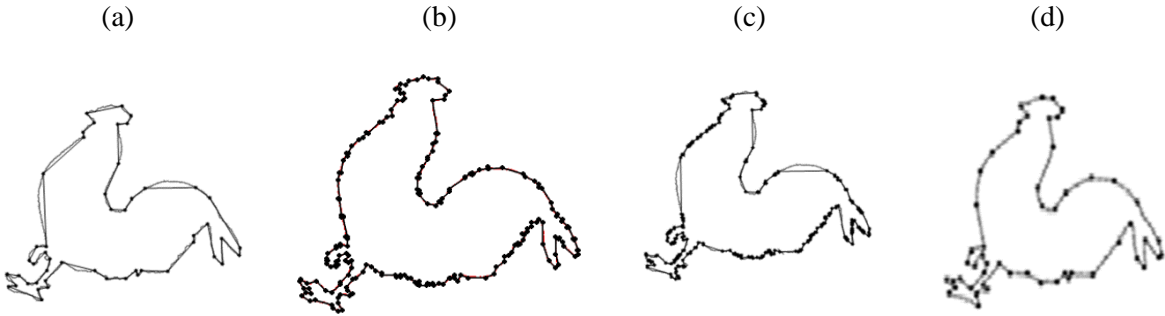

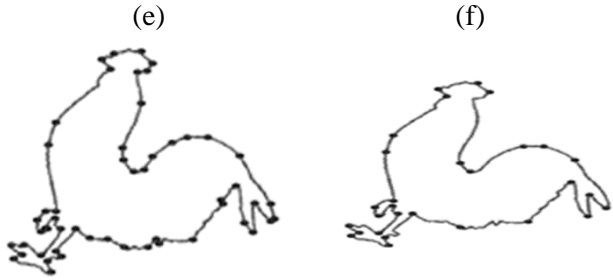

**Fig. 11:** Final approximation of chicken-5 contour by various method a) Carmona [11] at 54 DPs b)Prasad[23]RDP_opt at 218 DPs c) Prasad [23]Carmona_opt 258 DPs d) Fernandez [25] at 54 DPs e) proposed method at 54 DPs f)Proposed method at 29 DPs.

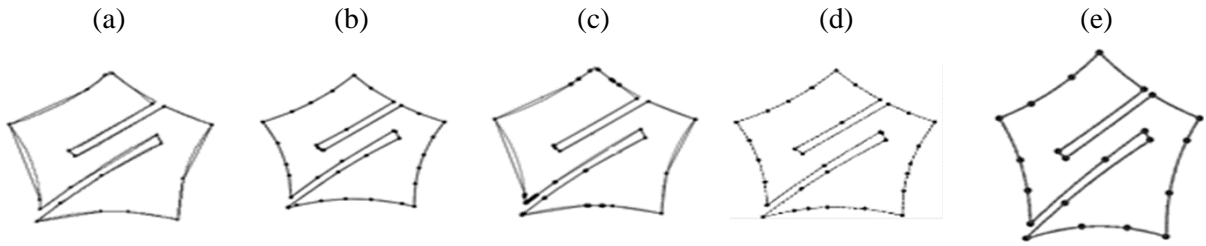

**Fig.12:** Final approximation obtained from device6-9 curve a) Carmona [11] at 22 DPs b)Prasad [23] RDP_opt at 38 DPs c) Prasad[23] Carmona_opt at 77 DPs d) Fernandez [25] at 33 DPs e) Proposed method at 22 DPs

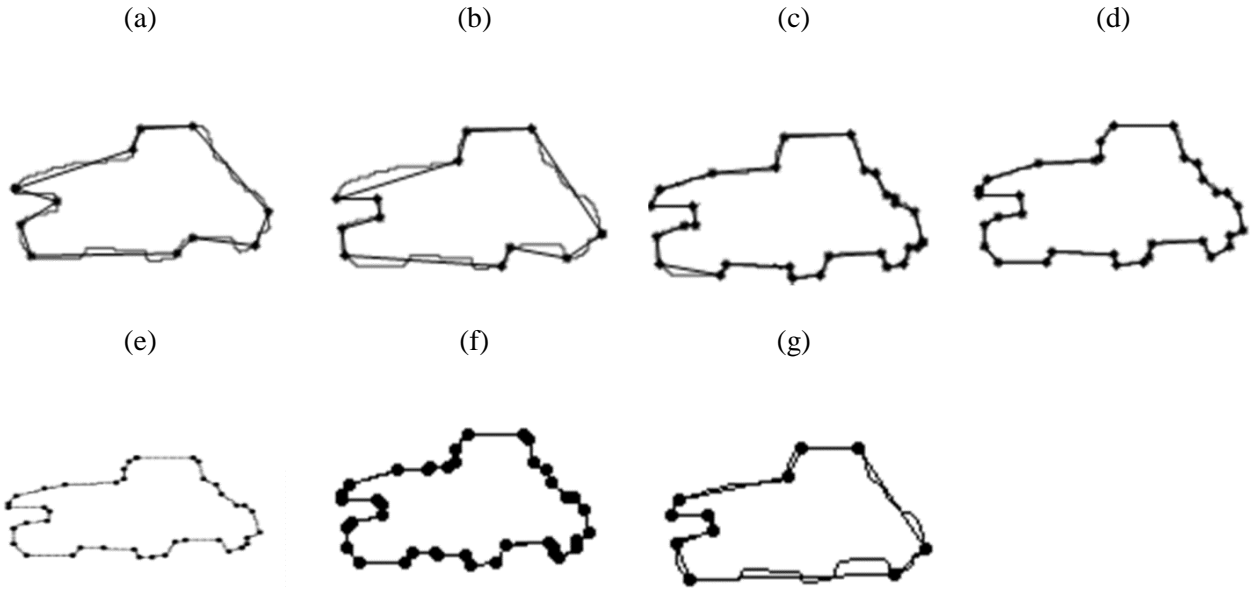

**Fig.13:** Final approximation obtained from truck-07 curve a)Masood[9] at 11 DPs b)Carmona [11] at 12 DPs, c) Prasad[23] Carmona_opt at 29 DPs d)Prasad [23] RDP_opt at 33 DPs e) e) Fernandez [25] at 40 DPs f) Proposed method at 44DPs g)Proposed method at 11 DPs.

### 4.1.3 Rotation Invariance

To test the efficiency of the proposed method against rotation invariance, bell-7 contour is rotated using varying amount angle. Then the rotated contour is given as an input to the proposed method as well as to the technique in [9]. The results are summarized for the reader's perusal. How to measure a technique is rotation invariant or to what extent? The answer is the metrics such as area of polygon, perimeter and compactness may be suggested to use along with results from human perception. The authors in [41] use the above mentioned metrics to prove whether the technique is able to produce the polygon with same positioned points before as well as after the rotation. This can be measured using compactness metric. Moreover the authors in [41] demonstrated the techniques proposed in a [9,11,12] are scaling as well as translation invariant using compactness metric.

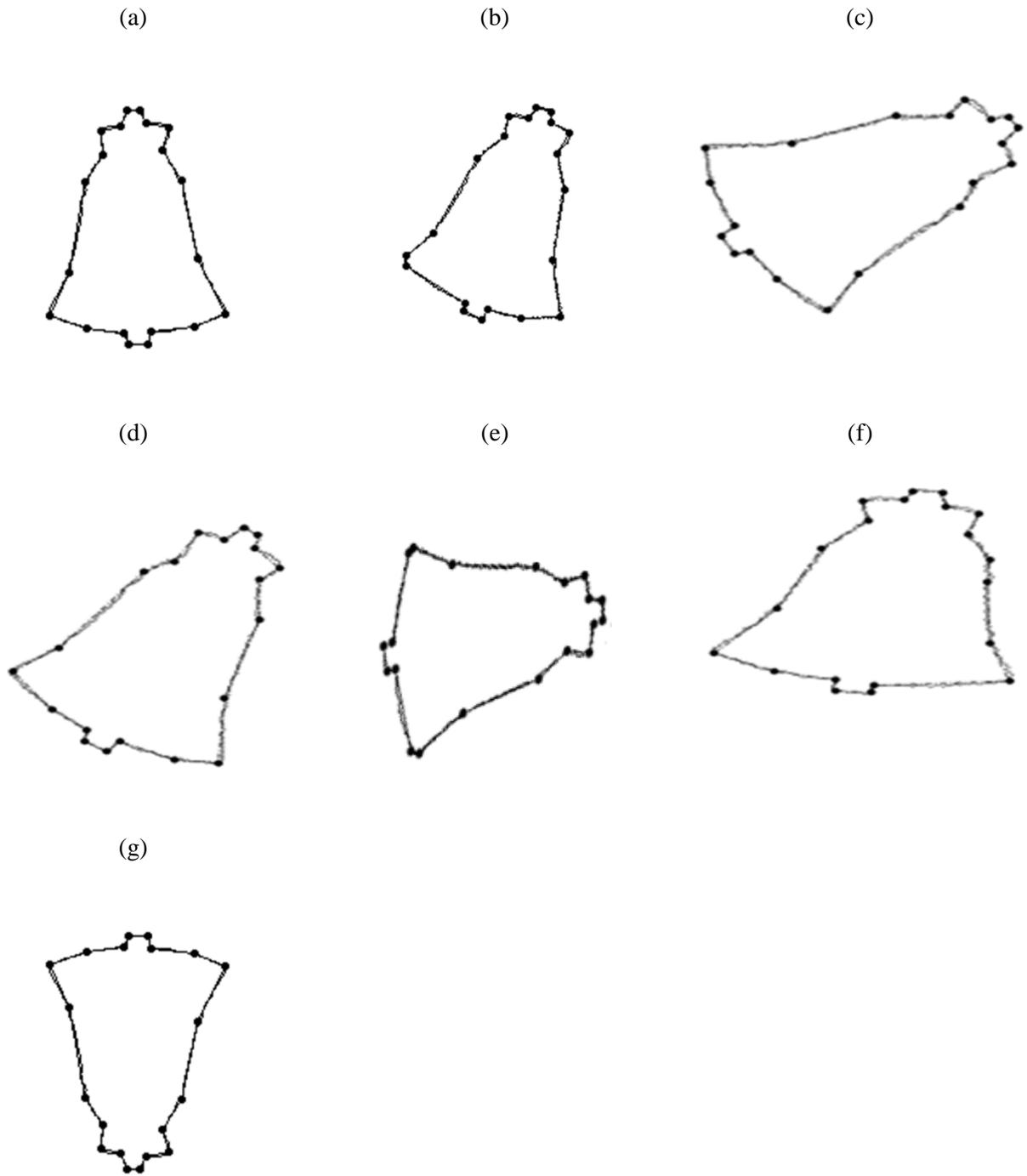

**Fig.14: The output polygon at 20 DPs by proposed methods in varying amount of angles a)Polygon at 20 DPs b)Polygon at 20°
c)Polygon at 30°  d) Polygon at 45° e) Polygon at 70° f) Polygon at 80° g) Polygon at 180°**

The mathematical interpretation of compactness metric (*COMP*) has been mentioned in eq(2). The Table 4 summarizes the value obtained by using COMP for the bell-7 contour by the proposed method.

$$comp = Area/Perimeter^2 \qquad (4)$$

**Table 4. Robustness of the proposed method against rotation using quantitative measurement**

| Contour | k | max($d_m$) | *ISE* | *Area* | *Perimeter* | *Compactness* |
|---|---|---|---|---|---|---|
| Bell-7 | 20 | 2.03 | 210.164 | 9231 | 299.13 | 0.10 |
| Bell-7 at 20° | | 2.60 | 281.57 | 9.2475e+03 | 343.81 | 0.07 |
| Bell-7 at 30° | | 2.70 | 348.868 | 9260 | 358.08 | 0.07 |
| Bell-7 at 45° | | * | * | 9258 | 362.32 | 0.07 |
| Bell-7 at 70° | | 2.91 | 325.29 | 9.254.5e+03 | 344.83 | 0.07 |
| Bell-7 at 80° | | 2.59 | 319.50 | 9151 | 327.10 | 0.08 |
| Bell-7 at 180° | | 2.03 | 210.164 | 9231 | 299.13 | 0.10 |

**Table 5. Robustness of the Masood[9] against rotation using quantitative measurement**

| Contour | k | max($d_m$) | *ISE* | *Area* | *Perimeter* | *Compactness* |
|---|---|---|---|---|---|---|
| Bell-7 | 20 | 3.48 | 315.00 | 6835 | 321.78 | 0.06 |
| Bell-7 at 20° | | 2.77 | 311.84 | 9130 | 333.32 | 0.08 |
| Bell-7 at 30° | | 1.99 | 270.51 | 9.1255e+05 | 190.61 | 0.25 |
| Bell-7 at 45° | | 2.02 | 257.51 | 9.14005e+03 | 366.72 | 0.06 |
| Bell-7 at 70° | | 3.79 | 381.35 | 9.1615e+03 | 343.59 | 0.07 |
| Bell-7 at 80° | | 2 | 266.677 | 9.1585e+03 | 326.90 | 0.08 |
| Bell-7 at 180° | | 3.487 | 315.00 | 6835 | 306.95 | 0.07 |

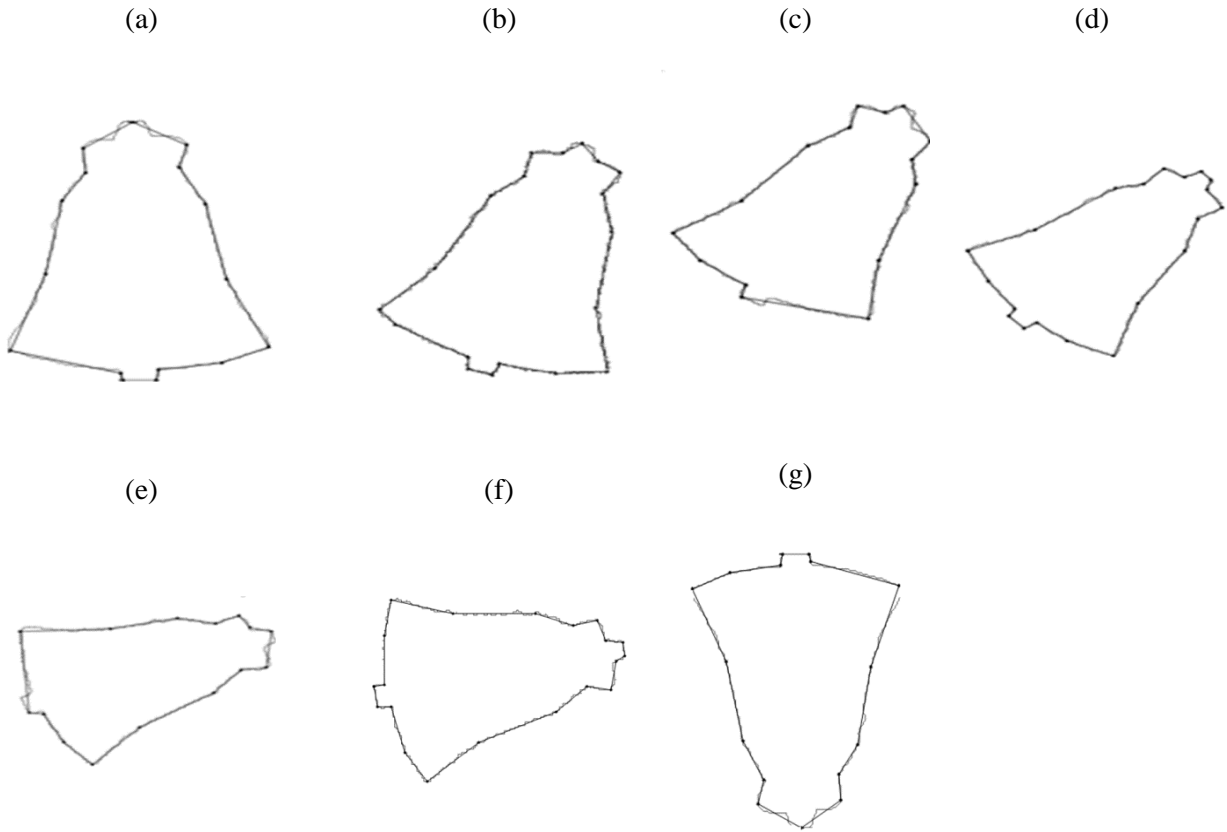

Fig.15: The output polygon at 20 DPs by Masood [9] in varying amount of angles a)polygon at 20 DPs b)Polygon at 20° c)Polygon at 30° d) Polygon at 45° e) Polygon at 70° f) Polygon at 80° g) Polygon at 180°

To compare the robustness of the technique against rotation the snapshots using bell-7 contour are displayed in Fig. 14 and Fig. 15. The output polygon at 20 amounts of dominant points is used here to check the technique is robust enough against rotation invariance. Most of the techniques considered in this paper produces polygon in non-parametric mode. The best thing to compare the efficiency of rotation invariance is to compare the output at minimal possible amount of points since the input curve may contain more redundant points. So the result of the proposed method is compared with Masood [9]. By using [9] any researcher can produce curve with specified number of dominant points. In Table 4 the value for geometric invariance assessment metrics (area of polygon, perimeter and compactness) reveals that results by proposed method using rotated contours measure against compactness metric is more or less nearer to the value produced by the proposed method before rotation and the visual snapshots in Fig. 14 are also supports the same. The results of Masood [9] in terms of quantitative measurements can be found in Table 5. Bell-7 at 30° value for compactness metric is varies high while comparing the results obtained at before rotation. In the remaining angles the rotated contours compactness metric more or less nearer to the value obtained by the method before rotation. Masood [9] snapshots can be found in Fig. 15. When we observe the position of dominant points on the output curve preserved by the Masood [9] are drastically in different position in after rotated curve approximation. Whereas the proposed methods try to maintain the same positioned dominant points in the rotated contours too see Fig. 14.

## 5 CONCLUSION

The proposed significant measure computing metric predicts the position of a projection of an every boundary point between its candidate line-segment thereby invokes suitable significant measure computing metric and accumulates their significant measure to define the significant value of every candidate of dominant points. The technique is demonstrated using wide variety of data sets, where the image contours are with different level details in terms of curvature as well as size. The proposed technique suits for any computer vision application desire to produce the digital boundary with minimal number points without compromising its shape according to human perception as well as using bench-marking performance measuring metrics.